\documentclass[runningheads]{llncs}
\usepackage[T1]{fontenc}
\usepackage{graphicx}
\usepackage{amsmath}
\usepackage{tabularray}
\usepackage{adjustbox}
\usepackage{multirow}
\usepackage{booktabs}
\usepackage{amssymb}
\usepackage{tikz}
\usepackage{xcolor}
\usepackage{colortbl}
\usepackage{float}
\usepackage{enumitem}
\newcommand{\para}[1]
{\vspace{.05in}\noindent\textbf{#1}}
\usepackage{hyperref}
\usepackage{marvosym}

\begin{document}
\title{Toward Medical Deepfake Detection: A Comprehensive Dataset and Novel Method}
\titlerunning{Toward Medical Deepfake Detection}
%
\author{Shuaibo Li\inst{1} \and
Zhaohu Xing\inst{1} \and
Hongqiu Wang\inst{1} \and
Pengfei Hao\inst{1} \and
Xingyu Li\inst{1} \and \\
Zekai Liu\inst{1} \and
Lei Zhu\inst{1,2} \textsuperscript{(\Letter)}}
\authorrunning{S. Li et al.}
%
\institute{The Hong Kong University of Science and Technology (Guangzhou) \and
The Hong Kong University of Science and Technology \\
\email{leizhu@ust.hk}\\
}



\maketitle              
\begin{abstract}
The rapid advancement of generative AI in medical imaging has introduced both significant opportunities and serious challenges, especially the risk that fake medical images could undermine healthcare systems. These synthetic images pose serious risks, such as diagnostic deception, financial fraud, and misinformation. However, research on medical forensics to counter these threats remains limited, and there is a critical lack of comprehensive datasets specifically tailored for this field. Additionally, existing media forensic methods, which are primarily designed for natural or facial images, are inadequate for capturing the distinct characteristics and subtle artifacts of AI-generated medical images. To tackle these challenges, we introduce \textbf{MedForensics}, a large-scale medical forensics dataset encompassing six medical modalities and twelve state-of-the-art medical generative models.  We also propose \textbf{DSKI}, a novel \textbf{D}ual-\textbf{S}tage \textbf{K}nowledge \textbf{I}nfusing detector that constructs a vision-language feature space tailored for the detection of AI-generated medical images. DSKI comprises two core components: 1) a cross-domain fine-trace adapter (CDFA) for extracting subtle forgery clues from both spatial and noise domains during training, and 2) a medical forensic retrieval module (MFRM) that boosts detection accuracy through few-shot retrieval during testing. Experimental results demonstrate that DSKI significantly outperforms both existing methods and human experts, achieving superior accuracy across multiple medical modalities.

\keywords{Medical Forensics \and Medical Image Analysis \and Trustworthy AI in Medical Imaging \and Healthcare Security \and Dataset}

\end{abstract}
\section{Introduction}
The rapid advancement of deep learning accelerated the integration of artificial intelligence (AI) in medicine. However, challenges such as the high cost of medical data collection, privacy regulations, and limited availability of annotated medical datasets have hindered progress. Recently, generative AI techniques, especially denoising diffusion models, have been utilized to generate realistic, diverse, and true-to-distribution medical imaging data, to augment healthcare model training \cite{31,32,33}. These AI-generated datasets have been beneficial in improving tasks like classification, segmentation, and cross-modal translation. Despite these advancements, the increasing quality and volume of synthetic medical images pose significant risks to healthcare systems. Prior studies \cite{12,13} have demonstrated that even medical experts can be misled by the high visual realism of AI-generated images. Thus, developing reliable and effective methods to detect these AI-generated medical images is crucial for mitigating risks and ensuring patient safety.

While AI-generated image detection has been widely studied in natural and facial images~\cite{43,44,45,46,li2021image,li2024unionformer}, few studies have focused on medical deepfakes. Applying methods intended for natural or facial images to the medical domain is suboptimal due to the unique nature of medical images. AI-generated medical images aim to replicate physiological phenomena and anatomical structures \cite{31,32,33,xing2024segmamba,wang2024video}. Compared to natural or facial images, these forged medical images often contain more subtle and localized clues (e.g., irregular low-level textures, unrealistic anatomical and pixel statistics), which are less semantically meaningful. Additionally, the variety of modalities and structures in medical images makes detection even more difficult. Current medical forensic methods \cite{48,49} are still in the early stage, primarily addressing images manipulated by traditional tools or GANs, but remain ineffective against the hyper-realistic images produced by advanced AI models like Diffusion Models. In this paper, we focus on detecting AI-generated medical images created by various state-of-the-art medical generative models. 

A major challenge in medical forensics is the lack of large-scale datasets of AI-generated medical images, due to the diversity of medical modalities and the complexity of generative models. To address this, we introduce MedForensics, a large-scale dataset of high-quality medical images generated by twelve leading models across six modalities, providing a key benchmark for medical forensics. Additionally, we propose the Dual-Stage Knowledge Infusing Detector (DSKI), designed to distinguish AI-generated medical images. In the training stage, a cross-domain fine-trace adapter (CDFA) captures forensic clues in the spatial and noise domains, using an inception module to extract multi-scale artifacts and a constrained CNN to model low-level pixel statistics. In the testing stage, a Medical Forensic Retrieval Module (MFRM) enhances detection performance and scalability. Experimental results show that DSKI outperforms state-of-the-art methods and human experts in detecting medical deepfakes, offering a crucial solution to mitigate the risks posed by AI-generated images and safeguard healthcare systems.

\begin{table*}[!t]
    \centering
    \caption{Details of our proposed MedForensics dataset.}
     \label{table1}
    \small
    \setlength{\tabcolsep}{5pt}
    \begin{adjustbox}{valign=c, width=0.9\textwidth}
    \begin{tabular}{ccccc}
        \toprule
        Model           & Modality       & Classification & Pairs Number & Real Image Source \\
        \midrule
        U-KAN-U         & Ultrasound     & Breast~        & 5000         & BUSI,UNS \\
  U-KAN-E        & Endoscope      & Colorectum     & 5000         & Kvasir, CVC, CV3D, ETIS \\
        U-KAN-H         & Histopathology & Histopathology & 5000         & GlaS, LC \\
        SegGuidedDiff-M & MRI            & Breast         & 5000         & DBCM \\
        SegGuidedDiff-C & CT             & Neck-to-pelvis & 5000         & CT-ORG \\
        Cheff          & X-ray          & Chest          & 5000         & MaCHeX \\
        EFN            & Ultrasound     & Heart          & 4000         & CAMUS \\
        MONAI-MGD       & MRI            & Brain          & 5000         & BraTS \\
        DiffEcho       & Ultrasound     & Heart          & 4000         & CAMUS \\
        ArSDM          & Endoscope      & Colorectum     & 5000         & Kvasir, CVC, CV3D, ETIS \\
        Hyper-GAN       & MRI            & Brain          & 5000         & BraTS \\
        ICVAE      & MRI            & Brain          & 5000         & BraTS \\
        
        \bottomrule
    \end{tabular}
    \end{adjustbox}
\end{table*}

\section{The Proposed MedForensics Dataset}
\subsection{Dataset Details}
\begin{figure*}[t]
\centering
\includegraphics[width=0.95\textwidth]{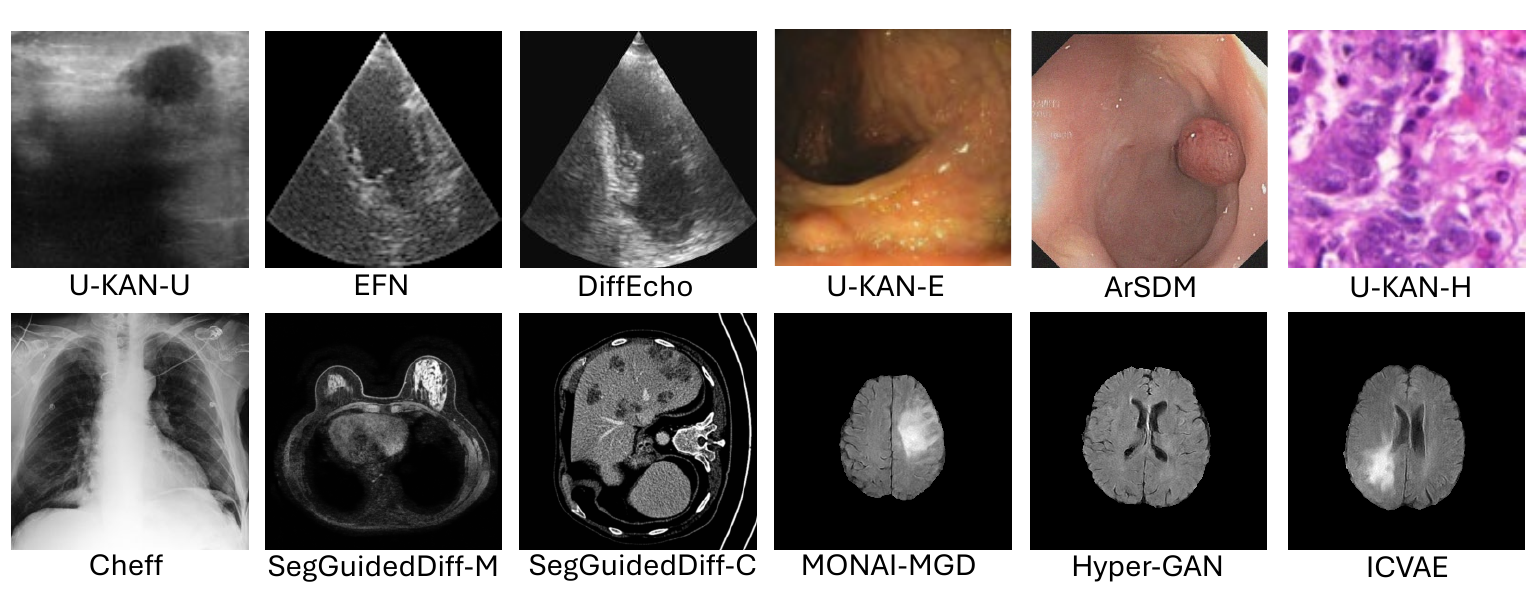}
\caption{Visualization of AI-generated medical images from the proposed MedForensics dataset, covering six imaging modalities and twelve generative models.} \label{fig2}
\end{figure*}

To advance the development of medical forensic detectors and assess their ability to distinguish AI-generated from real medical images, we introduce MedForensics, a large-scale dataset comprising 116,000 medical images, with an equal number of real and AI-generated fake images. MedForensics spans six medical modalities: Ultrasound, Endoscopy, Histopathology, MRI, CT, and X-ray, covering a diverse range of real-world forensic scenarios. We employ 12 state-of-the-art (SOTA) models from nine medical image generation studies, including U-KAN~\cite{31}, SegGuidedDiff~\cite{32}, Cheff~\cite{33}, EFN~\cite{34}, MONAI-MGD~\cite{35}, DiffEcho \cite{36}, ArSDM \cite{37}, Hyper-GAN \cite{38}, and ICVAE \cite{39}. Each generator produces 5,000 synthetic images, except for DiffEcho and EFN, which produce 4,000 images. Real images are primarily sourced from the corresponding generative model's training set to maintain a balanced distribution of real and fake images. In cases where training sets were small 
(e.g., U-KAN-H's GlaS dataset with only 165 images), we supplement with images from larger datasets of the same modality \cite{cv3d,LC,UNS,kvasir,33,camus}. The dataset is standardized at the 256×256 resolution, and the images are split into training and testing sets with an 80/20 split. With its large volume, diverse modalities, and inclusion of advanced SOTA models, MedForensics provides a comprehensive resource for developing and evaluating medical forensic detection methods. Table~\ref{table1} details the dataset composition, while Figure~\ref{fig2} illustrates examples of the generated images.

\subsection{Fake Medical Image Generators}
\textbf{Diffusion-based Model:} Diffusion models have recently emerged as the most advanced architecture in medical image generation. Notably,  U-KAN~\cite{31} integrates Kolmogorov-Arnold Networks (KANs) into the noise predictor of the diffusion model, with three variants that generate highly realistic ultrasound, endoscopy, and histopathology images, namely U-KAN-U, U-KAN-E, and U-KAN-H. SegGuidedDiff~\cite{32} enables anatomically controllable image generation by adhering to a multi-class anatomical segmentation mask during each sampling step. SegGuidedDiff-M and SegGuidedDiff-C generate high-quality breast MRI and abdominal/neck-to-pelvis CT images. Cheff~\cite{33} utilizes a cascaded latent diffusion model to generate state-of-the-art chest radiographs, while EFN~\cite{34} employs Denoising Diffusion Probabilistic Models (DDPM) guided by cardiac semantic labels to generate high-quality ultrasound images. MONAI-MGD, part of the widely used MONAI~\cite{35} library, generates brain MRI images. DiffEcho~\cite{36} produces realistic echocardiography images, and ArSDM~\cite{37} applies an adaptive refinement semantic diffusion model to generate colonoscopy images. 

\para{GAN-based Model:} Over the past decade, Generative Adversarial Networks (GANs) have led to substantial advancements in the quality of image generation. Hyper-GAN~\cite{38} constructs a multi-contrast MRI image translation model that adapts to various MR contrast types, enabling high-fidelity brain MRI generation. 

\para{VAE-based Model:} Variational Autoencoders (VAEs) learn a latent space to encode and reconstruct data. ICVAE~\cite{39} separates the input data's embedding space from conditioning variables, ensuring that generated brain MRI features are independent of conditioning factors, thus increasing output diversity.  

\section{Method}
\subsection{Overview of DSKI}
\begin{figure*}[!t]
\centering
\includegraphics[width=0.95\textwidth]{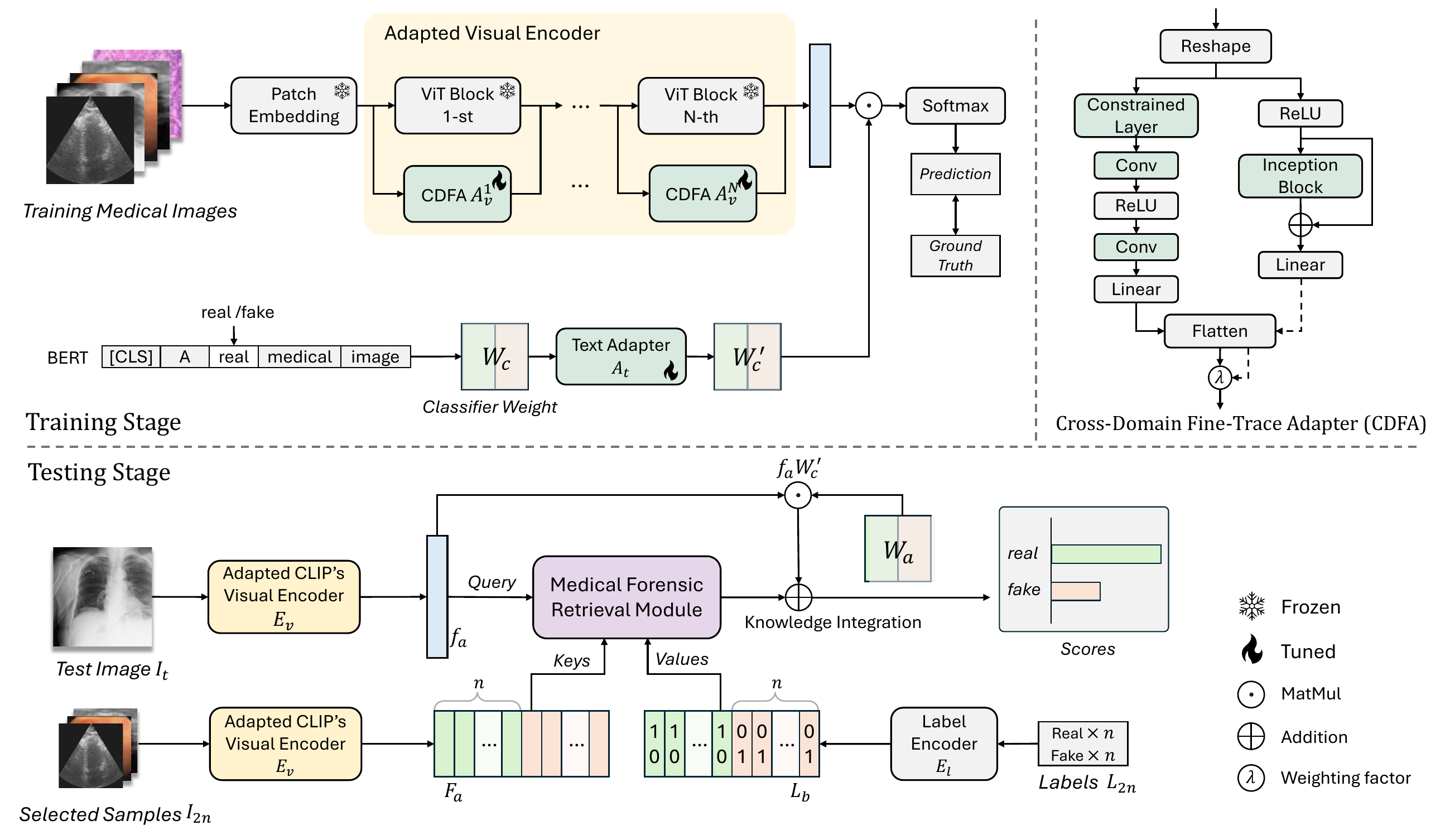}
\caption{Overview of the DSKI framework. The training stage employs a Cross-Domain Fine-Trace Adapter (CDFA) to inject medical forensic knowledge into the CLIP backbone via spatial and noise feature streams. During testing, a Medical Forensic Retrieval Module (MFRM) retrieves few-shot knowledge from a feature bank to enhance detection performance. } \label{fig1}
\end{figure*}
As shown in Figure \ref{fig1}, DSKI is built upon the pre-trained CLIP \cite{21}, which has been trained on an extensive dataset of 400 million image-text pairs, providing a robust vision-language feature space suitable for a variety of downstream tasks. CLIP has also demonstrated sufficient effectiveness in media forensics tasks \cite{43},  which makes it an ideal foundation for our approach. While the original CLIP space performs well in distinguishing the authenticity of natural images, its high-level visual semantics are insufficient for medical image forensics due to the lack of fine-grained details and medical-specific information. To address this, DSKI performs a two-stage medical knowledge infusing to adapt the original CLIP feature space into an adequate, generalized medical forgery discrimination space. The training stage is centered around the cross-domain fine-trace adapter (Section \ref{3.2}), while the testing phase utilizes the forensic retrieval module (Section \ref{3.3}) to enhance detection accuracy further.

\subsection{Training Stage: Cross-domain Fine-trace Adapter (CDFA) }
\label{3.2}
Our training set includes real and fake medical images, each paired with natural language prompts (e.g., "A real/fake medical image").  The vanilla CLIP  employs the text encoder to convert the prompts $P_c$ into classifier weights $W_c$, and the visual encoder extracts image features $f$. We introduce visual adapters $A_v^{(i)}(\cdot)$ and a text adapter $A_t(\cdot)$ to fine-tune the pre-trained CLIP.  Each visual adapter (e.g., CDFA) is placed in parallel with the MLP in the $i$-th transformer block. The CDFA has two streams that capture fine-grained forgery traces from both the spatial and noise domains. In the noise domain, constrained convolution layers and ReLU layer help learn abnormal pixel relationships while suppressing semantic content. In the spatial domain, an inception module with parallel convolutions (1×1, 3×3, 5×5) captures multi-scale forensic artifacts, which are crucial for interpreting medical images \cite{li2024unionformer}.  The two streams are defined as follows:
\begin{equation}
\hat{f}_s^{(i)} = \operatorname{Conv} \big( \operatorname{ReLU} \big( \operatorname{ConstrainedConv} \big( f^{(i)} \big) \big) \big),
\end{equation}
\begin{equation}
\hat{f}_n^{(i)} = \operatorname{InceptionConv} \big( \operatorname{ReLU} \big( f^{(i)} \big) \big),
\end{equation}
where $f^{(i)}$ is the input to the visual adapter in the $i$-th transformer block, and $\hat{f}_s^{(i)}$ , $\hat{f}_n^{(i)}$ are the outputs from the spatial and noise streams.  The outputs of both streams are fused using a learnable scale factor $\lambda$, forming the final medical forensic feature: 
\begin{equation}
\hat{f}^{(i)}=\hat{f}_s^{(i)}+\lambda \hat{f}_n^{(i)}. 
\end{equation}
The adapter output $\hat{f}^{(i)}$ is added to the corresponding MLP. The text adapter $A_t(\cdot)$ consists of two sequential linear layers applied after the text encoder. At this point, we obtain the new image feature $f^{\prime}$ and classifier weights $W^{\prime}$.  During training, the weights of $A_v(\cdot)$ and $A_t(\cdot)$ are optimized using the binary cross-entropy loss:  
\begin{equation}
    \mathcal{L}(\theta)=-\frac{1}{N} \sum_{n=1}^N\left[y_n \log \left(p_n\right)+\left(1-y_n\right) \log \left(1-p_n\right)\right],
\end{equation}
where $N$ is the total number of training samples, $y_n$ is the ground truth label for the $n$-th sample (0 for real, 1 for fake), and $p_n$ is the predicted probability of the sample being classified as the positive class.
\subsection{Testing Stage: Medical Forensic Retrieval Module (MFRM)}
\label{3.3}
After the training stage, DSKI collects  cross-domain, multi-scale artifacts to form a comprehensive fine-grained view for medical image forensics. To enhance detection robustness and accuracy, we design MFRM, inspired by \cite{29}. The MFRM is built upon a cache feature bank. Using the fully adapted CLIP model from the training stage, we randomly select a small sample set $I_{2n}$ from the completed dataset, containing $n$ real and $n$ synthetic medical images with corresponding labels $L_{2n}$. The adapted visual encoder $E_v$ extracts the L2-normalized features $F_a \in \mathbb{R}^{2 n \times C}$ . The ground-truth labels $L_{2n}$ are converted into a binary vector $L_b \in \mathbb{R}^{2 n \times 2}$ with a one-hot encoder $E_l$. $F_a$ and $L_b$ serve as keys and values, forming the feature bank, which stores the new medical forensic knowledge extracted from the few-shot samples.  
Given a test image $I_t$, its L2-normalized feature $f_a \in \mathbb{R}^{1\times C}$ is extracted by the adapted CLIP visual encoder and is used as a query in the MFRM to retrieve relevant information from the feature bank.  The affinity scores $A_s \in \mathbb{R}^{1\times 2n}$ between the query and the keys are computed as:
\begin{equation}
    A_s=\exp \left(-\alpha\left(1-f_a {F}_a^T\right)\right),
\end{equation}
 where $\alpha$  is a modulation hyperparameter. The prediction of the MFRM can be obtained with $A_sL_b$. Then, the knowledge retrieved from the MFRM is combined with prior knowledge ($f_a W_a^T \in \mathbb{R}^{1 \times 2}$, where $W_a$ is the adapted classifier weights) from the adapted CLIP model to yield the final logits:  
 \begin{equation}
     p_f=\beta A_s L_b+f_a W_a^{T},
 \end{equation}
where $\beta$ is the residual ratio. This retrieval-based MFRM not only enhances synthetic medical image detection by integrating new knowledge without retraining, but also provides scalability for our model.  As new medical image synthesis frameworks emerge, DSKI can  seamlessly integrate few-shot, newly labeled samples into the feature bank during testing, boosting its ability to detect  fake images from new frameworks.
\section{Experiments}
\subsection{Experiments Setup}
We evaluated the performance of our proposed method on the MedForensics (see Section 2). The compared methods fall into three categories: state-of-the-art natural image forensics (UniFD \cite{43}, AEROBLADE \cite{44}, NPR \cite{45}), facial image forensics (F3-Net \cite{46}, RECCE \cite{47}), and medical image forensics (MedNet \cite{48}, DFH \cite{49}). All compared methods were trained on the MedForensics training set for fair comparison. We evaluated all models on the MedForensics dataset using Accuracy (Acc) and Average Precision (AP), following standard media forensics protocols.

\para{Implementation Details.}
We initialize the backbone (ViT-L/14) using weights provided by \cite{21}, and the remaining parts are randomly initialized. We train the model for 50 epochs with a batch size of 32 on 4 NVIDIA GTX3090 GPUs. We adopt SGD as the optimizer, with a momentum of 0.9, a weight decay of 0.005, and an initial learning rate of 1e-4. Meanwhile, the visual adapters are placed in the ViT blocks at $i = 7, 15, 23$. $\alpha$ and $\beta$ are set to 0.1 and 10 for blending adjustment. In testing, the number of selected samples $n$ is empirically set to 16.

\subsection{Comparison Results}
\textbf{Quantitative Results Comparison.} Table \ref{table_method} summarizes the detection performance of different methods across various modalities.   Thanks to the dual-stage aggregation of medical forensic knowledge, DSKI significantly outperforms other methods in all modalities and metrics. Specifically, DSKI shows an improvement of over 20\% in Acc and AP compared to UniFD \cite{43} using the original CLIP feature space. Unlike methods \cite{48,49}, which suffer degraded performance in detecting fake images (with relatively poor quality) from outdated GANs, DSKI handles a broader range of architectures and outperforms these methods by a significant margin.

\begin{table*}[!t]
    \centering
    \caption{Cross-modal detection Acc/AP of compared methods on the MedForensics dataset. The bold
value indicates the best performance.}
    \small
    \setlength{\tabcolsep}{0pt}
    \begin{adjustbox}{valign=c, width=\textwidth}
    \begin{tabular}{
        >{\centering\arraybackslash}p{2.7cm}
        *{7}{>{\centering\arraybackslash}p{1.9cm}}
    }
        \toprule
        Methods & Ultrasound & Endoscope & Histopathology & MRI & CT & X-ray & Mean \\
        \midrule
        UniFD \cite{43} & 85.7/78.8 & 80.8/86.7 & 84.2/78.2 & 61.6/54.5 & 54.6/66.1 & 63.6/67.4 & 71.8/72.0 \\
        AEROBLAD \cite{44} & 73.1/65.8 & 68.7/73.9 & 55.6/51.4 & 64.8/81.6 & 56.3/56.8 & 57.9/56.6 & 62.7/64.4 \\
        NPR \cite{45} & 71.8/76.5 & 56.2/58.4 & 75.4/74.7 & 51.5/56.1 & 74.7/73.3 & 52.2/63.7 & 63.6/67.1 \\
        F3-Net \cite{46} & 72.3/77.2 & 70.8/71.5 & 59.8/64.3 & 55.9/57.8 & 65.6/72.3 & 53.5/58.0 & 63.0/66.9 \\
        DFH \cite{49} & 72.8/73.5 & 62.4/68.3 & 75.2/72.4 & 59.3/58.3 & 67.2/70.3 & 64.8/64.7 & 67.0/67.9 \\
        RECCE \cite{47} & 59.8/62.3 & 54.7/61.0 & 64.6/69.7 & 69.0/69.5 & 51.4/62.8 & 63.3/85.1 & 60.5/68.4 \\
        MedNet \cite{48} & 63.2/60.7 & 58.3/54.5 & 64.3/62.3 & 70.9/68.7 & 69.7/69.4 & 64.4/66.7 & 65.1/63.7 \\
         \rowcolor{green!10} \textbf{Ours} & \textbf{98.9/93.4} & \textbf{91.9/91.4} & \textbf{97.8/93.8} & \textbf{84.9/89.1} & \textbf{92.7/93.6} & \textbf{83.4/91.6} & \textbf{91.6/92.2} \\
        \bottomrule
    \end{tabular}
    \end{adjustbox} 
    \label{table_method}
\end{table*}

\para{Turing Test (Blinded Expert Evaluation).} To evaluate the real-world applicability of our method, we conducted a Turing test with three medical specialists—a radiologist, a pathologist, and a gastroenterologist. We randomly selected 50 synthetic and 50 real images from each modality in the MedForensics dataset and asked the experts to assess their authenticity. Table \ref{table_Turing} shows that the experts had difficulty distinguishing real from fake images due to the similarity in pathological features. In contrast, our DSKI effectively identified the fake images by capturing low-level medical forgery artifacts.

\begin{table*}[!t]
\centering
\caption{Blinded expert evaluation: human vs. DSKI performance.}
    \begin{adjustbox}{valign=c, width=0.86\textwidth}
\small
\setlength{\tabcolsep}{6pt}

\begin{tabular}{cccc}
\toprule
Detectors & Ultrasound, MR, CT, X-ray & Histopathology & Endoscope \\ 
\midrule
Radiologist      & 72.3/78.7                      &          -              &        -             \\ 
Pathologist         &                               -   & 74.1/78.9           &   -               \\ 
Gastroenterologist   & -                      & -             &      68.5/72.0     \\ 
 \rowcolor{green!10} DSKI                &\textbf{89.8/90.1}                     & \textbf{95.8/93.7}              & \textbf{92.0/90.7}          \\ 
\bottomrule
\end{tabular}
\end{adjustbox}
\label{table_Turing}
\end{table*}
\subsection{Ablation Study}
We conducted ablation experiments on the core components (Table \ref{ab_a}). Integrating both modules significantly enhances performance. Each component plays a distinct role: CDFA captures fine-grained artifacts during training, while MFRM utilizes a retrieval mechanism to inject medical forensic knowledge from few-shot samples during testing. Table \ref{ab_b} summarizes the evaluation of different CDFA feature streams, with the full configuration achieving optimal results. We also tested the DSKI's scalability on 1,000 fake UWF fundus images generated by an unseen framework \cite{412} and 1000 real images. DSKI demonstrated strong performance (Acc $=$ 86.6\%, AP $=$ 89.4\%) even without adding new samples. Adding a few extra samples to the feature bank improved Acc by 2.1\% and AP by 2.7\%, showcasing its scalability to emerging threats.

\begin{table*}[htbp]
    \centering
    \small
    \begin{adjustbox}{valign=c}
    \begin{tabular}{cc}
        
        \begin{minipage}[t]{0.48\textwidth}
            \centering
            \caption{Ablation study for core components (CDFA and MFRM) in DSKI.}
            \setlength{\tabcolsep}{6pt}
            \begin{tabular}{cccc}
                \toprule
                CDFA & MFRM & Acc & AP \\
                \midrule
                $\times$ & $\times$ & 70.9 & 72.6 \\
                $\checkmark$ & $\times$ & 86.0 & 89.8 \\
                $\times$ & $\checkmark$ & 75.3 & 78.9 \\
                \rowcolor{green!10} $\checkmark$ & $\checkmark$ & \textbf{91.6} & \textbf{92.2} \\
                \bottomrule
            \end{tabular}
            \label{ab_a}
        \end{minipage}
        &
       
        \begin{minipage}[t]{0.48\textwidth}
            \centering
            \caption{Ablation study for feature streams (spatial and noise) in CDFA.}
            \setlength{\tabcolsep}{6pt}
            \begin{tabular}{cccc}
                \toprule
                Spatial & Noise & Acc & AP \\
                \midrule
                $\times$ & $\times$ & 75.3 & 78.9 \\
                $\checkmark$ & $\times$ & 80.3 & 82.4 \\
                $\times$ & $\checkmark$ & 82.6 & 83.3 \\
                \rowcolor{green!10} $\checkmark$ & $\checkmark$ & \textbf{91.6} & \textbf{92.2} \\
                \bottomrule
            \end{tabular}
            \label{ab_b}
        \end{minipage}
    \end{tabular}
    \end{adjustbox}
\end{table*}

\section{Conclusion}
This paper tackles the growing threat of AI-generated medical deepfakes with two key contributions. First, we introduce MedForensics, a high-quality, large-scale dataset for medical forensics, covering six modalities and generated by twelve state-of-the-art models. Second, we propose the Dual-Stage Knowledge Infusing (DSKI) detector, a two-stage method that enhances synthetic medical image detection. Experimental results show that DSKI outperforms both state-of-the-art methods and human experts across multiple modalities, offering a more robust solution for detecting medical deepfakes. This work facilitates the development of trustworthy healthcare systems by providing a comprehensive dataset and an effective detection framework.

\begin{credits}
\subsubsection{\ackname}
This work is supported by the Guangdong Science and Technology Department (No. 2024ZDZX2004).

\subsubsection{\discintname}
The authors have no competing interests to declare that are relevant to the content of this article.
\end{credits}

\bibliographystyle{splncs04} 
\bibliography{Parper-2888}

\end{document}